\documentclass{article} % For LaTeX2e
\usepackage{iclr2023_conference,times}

\usepackage{amsmath,amsfonts,bm}

\def\eqref#1{equation~\ref{#1}}
\def\1{\bm{1}}

\DeclareMathAlphabet{\mathsfit}{\encodingdefault}{\sfdefault}{m}{sl}
\SetMathAlphabet{\mathsfit}{bold}{\encodingdefault}{\sfdefault}{bx}{n}

\usepackage{hyperref}
\usepackage{url}
\usepackage{multirow}
\usepackage{booktabs} % for professional tables
\usepackage{graphicx}
\usepackage{natbib} 
\usepackage{xcolor}         % colors
\usepackage{soul}
\usepackage{subcaption}
\title{Federated Learning for Inference at Anytime and Anywhere}

\iclrfinalcopy

\author{Zicheng Liu$^*$ \ \\
School of Information Science \\ 
\& Electronic Engineering \\
Zhejiang University\\
Hangzhou, China \\
\texttt{liuzicheng@westlake.edu.cn} \\
\And 
Da Li\thanks{indicates equal contribution.} \ \\
Machine Learning \& Data Intelligence \\
Samsung AI Center  \\
Cambridge, UK  \\
\texttt{dali.academic@gmail.com}
\And
Javier Fernandez-Marques \\
Automated AI \\
Samsung AI Center \\
Cambridge, UK \\
\texttt{j1.fernandez@samsung.com}
\And 
Stefanos Laskaridis \\
Distributed AI \qquad\qquad\qquad\qquad\quad~~ \\
Samsung AI Center \\
Cambridge, UK \\
\texttt{mail@stefanos.cc}
\And
Yan Gao \\
Department of Computer Science and Technology \\
The University of Cambridge \\
Cambridge, UK \\
\texttt{y.gaogy@gmail.com}
\And
Łukasz Dudziak \\
Automated AI \qquad\qquad\qquad\qquad\quad~~~~\\
Samsung AI Center \\
Cambridge, UK \\
\texttt{l.dudziak@samsung.com} 
\And
Stan Z. Li\thanks{indicates joint corresponding authors.} \ \\
School of Engineering\\
Westlake University\\
Hangzhou, China \\
\texttt{Stan.ZQ.Li@westlake.edu.cn} 
\And
    Shell Xu Hu$^\dagger$ \\
Machine Learning \& Data Intelligence~~ \\
Samsung AI Center \\
Cambridge, UK \\
\texttt{shell.hu@samsung.com} 
\And
Timothy Hospedales$^\dagger$ \\
Machine Learning \& Data Intelligence \\
Samsung AI Center \\
Cambridge, UK \\
\&
School of Informatics \\
The University of Edinburgh \\
Edinburgh, UK \\
\texttt{t.hospedales@samsung.com} 
}

\newcommand\Tstrut{\rule{0pt}{1.2ex}}         % = `top' strut
 
\newcommand{\alg}{Accumulator}

\definecolor{gray}{rgb}{0.5,0.5,0.5}
\definecolor{darkergreen}{RGB}{21, 152, 56}
\definecolor{gray94}{gray}{.94}
\definecolor{gray90}{gray}{.90}
\definecolor{gray85}{gray}{.85}

\newcommand{\gray}[1]{\textcolor{gray}{#1}}

\newif\ifcomment
\commenttrue
\ifcomment
    \newcommand{\steve}[1]{\sethlcolor{cyan}\hl{[Stefanos: #1]}}
    \newcommand{\tim}[1]{\sethlcolor{magenta}\hl{[Tim: #1]}}

\else
    \newcommand{\steve}[1]{}
    \newcommand{\tim}[1]{}
\fi
\begin{document}

\maketitle

%%%%%%%%%%%%%%% main %%%%%%%%%%%%%%%%

\begin{abstract}
Federated learning has been predominantly concerned with collaborative training of deep networks from scratch, and especially the many challenges that arise, such as communication cost, robustness to heterogeneous data, and support for diverse device capabilities. However, there is no unified framework that addresses all these problems together. 
% \steve{This is not true, we need to make this claim pretrained-specific.} 
This paper studies the challenges and opportunities of exploiting pre-trained Transformer models in FL. In particular, we propose to efficiently adapt such pre-trained models by injecting a novel attention-based adapter module at each transformer block that both modulates the forward pass and makes an early prediction. Training only the lightweight adapter by FL leads to fast and communication-efficient learning even in the presence of heterogeneous data and devices. Extensive experiments on standard FL benchmarks, including CIFAR-100, FEMNIST and SpeechCommandsv2 demonstrate that  this simple framework provides fast and accurate FL while supporting heterogenous device capabilities, efficient personalization, and scalable-cost anytime inference. 
% { \href{https://anonymous.4open.science/r/Federated-Learning-for-Inference-at-Anytime-and-Anywhere-356A/}{Our code can be found here}.}
\end{abstract}
\section{Introduction}
Federated learning (FL) was proposed by~\cite{mcmahan2017communication} as a new paradigm for distributed learning in which user data privacy is protected. Following the introduction of the FL setting, subsequent work focused on addressing the emerging challenges that arise due to FL constraints, 
%to the hard constraint on data privacy
such as communication cost~\cite{mishchenko2022proxskip}, data heterogeneity~\cite{li2020federated} and supporting diverse device hardware~\cite{horvath2021fjord,rapp2022distreal}. For example, to reduce the communication cost,  ideas borrowed from model compression, such as quantization \cite{alistarh2017qsgd, fu2020don}, sparsification \cite{stich2018sparsified} and pruning~\cite{yu2021adaptive,jiang2022model} have been successfully applied; to mitigate the non-IID issue of data heterogeneity, different model training recipes for optimization~\cite{li2020federated,wang2020tackling}, model initialization~\cite{nguyen2022begin} and architecture design~\cite{qu2022rethinking} have also been proposed. %Meanwhile, some works~\cite{horvath2021fjord,rapp2022distreal} were proposed to support different compute capabilities among clients.

% Specifically, during the collaborative learning in FL, each user client uses its own collected data to train a local model, which will be uploaded to the server after each round of local training without leaking the data information. 
% In such a way, a server is in charge of aggregating the trained models using some strategies, e.g. FedAvg~\cite{mcmahan2017communication} or FedAdam~\cite{reddi2020adaptive}, into a single model, which will be returned to clients. 
% This procedure will repeat until a stop criterion, e.g. the training model is converged~\cite{mcmahan2017communication}, is reached.

% proposed different optimization techniques FedProx~\cite{li2020federated} and FedNova~\cite{wang2020tackling}. And recently \cite{nguyen2022begin} found it useful to use pre-trained models, especially Transformers~\cite{qu2022rethinking}, to this end. Moreover, some works FjORD~\cite{horvath2021fjord} and DISTREAL~\cite{rapp2022distreal} were proposed to support the system heterogeneity, such as heterogeneous computational capacities, between different clients.

A new question has now emerged for FL community: Can we benefit from the recent success of large-scale centralized pre-training of foundation models \cite{bommasani2021opportunities}? 
Although contemporary federated learning has predominantly been concerned with collaborative training of deep models from scratch~\cite{mcmahan2017communication, li2020federated}, neglecting publicly available pre-trained models, it has been observed by ~\cite{qu2022rethinking} that fine-tuning pretrained vision transformers (ViT) significantly improves FL performance for various image recognition tasks and enables great robustness to the data heterogeneity among clients. 
Despite being an important step forward, fine-tuning the whole pre-trained ViT can be problematic due to the heavy communication cost of exchanging large numbers of model parameters and the weak capabilities of on-device training for many client devices. In this paper, we address this problem by reframing FL as a parameter-efficient (PE) downstream learning task. This is in line with the recent PE-based adaptation developments in centralized vision and natural language processing methods. 
%take a different perspective of treating FL as a downstream task of pre-trained foundation models (FMs) by parameter-efficient (PE) learning. PE learning has been of researchers' interest in various fields~\cite{rebuffi2017learning, mahabadi2021parameter, tomanek2021residual}, especially now in NLP with the soaring sizes of pretrained FMs~\cite{liu2019roberta, radford2019language, brown2020language}. People mainly focused on 
This line of parameter-efficient adaptation research includes adapters~\cite{rebuffi2017learning, houlsby2019parameter,tomanek2021residual}, prompt tuning~\cite{li2021prefix,lester2021power}, bias-only fine-tuning~\cite{zaken2021bitfit} and so on. 
We contribute a new adaptor suited for FL under the foundation model regime, which is designed for the requirements of adaptation to fit client devices at \emph{anytime} (under different compute and memory budgets) and \emph{anywhere} (under severe data heterogeneity). 

\begin{figure}[t]
     \centering
     \begin{subfigure}[t]{0.42\textwidth}
        \centering
        \includegraphics[width=\textwidth]{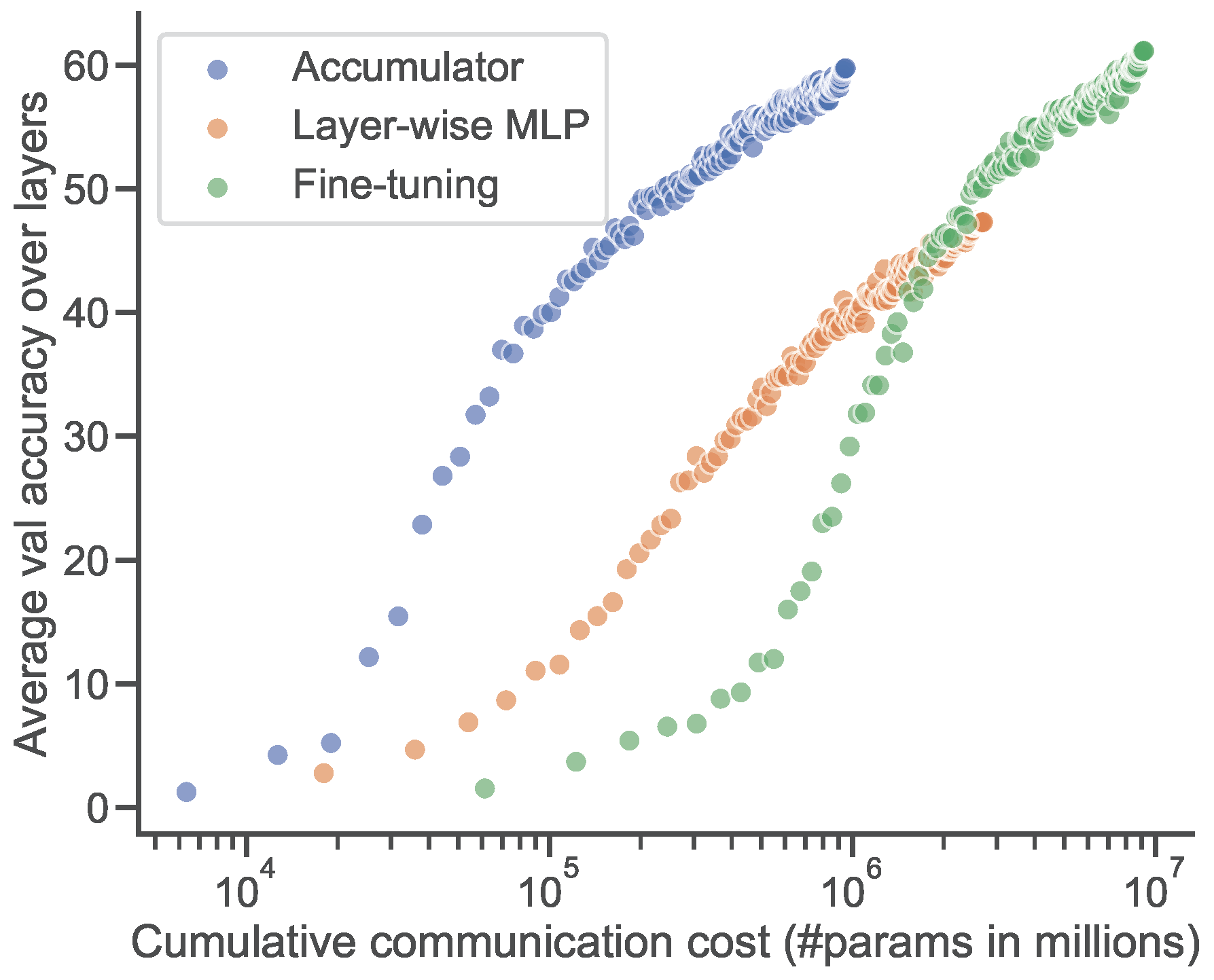}
        \caption{Non-IID case: $\alpha=0.1$}
    \end{subfigure}
     \qquad
    \begin{subfigure}[t]{0.42\textwidth}
        \centering
        \includegraphics[width=\textwidth]{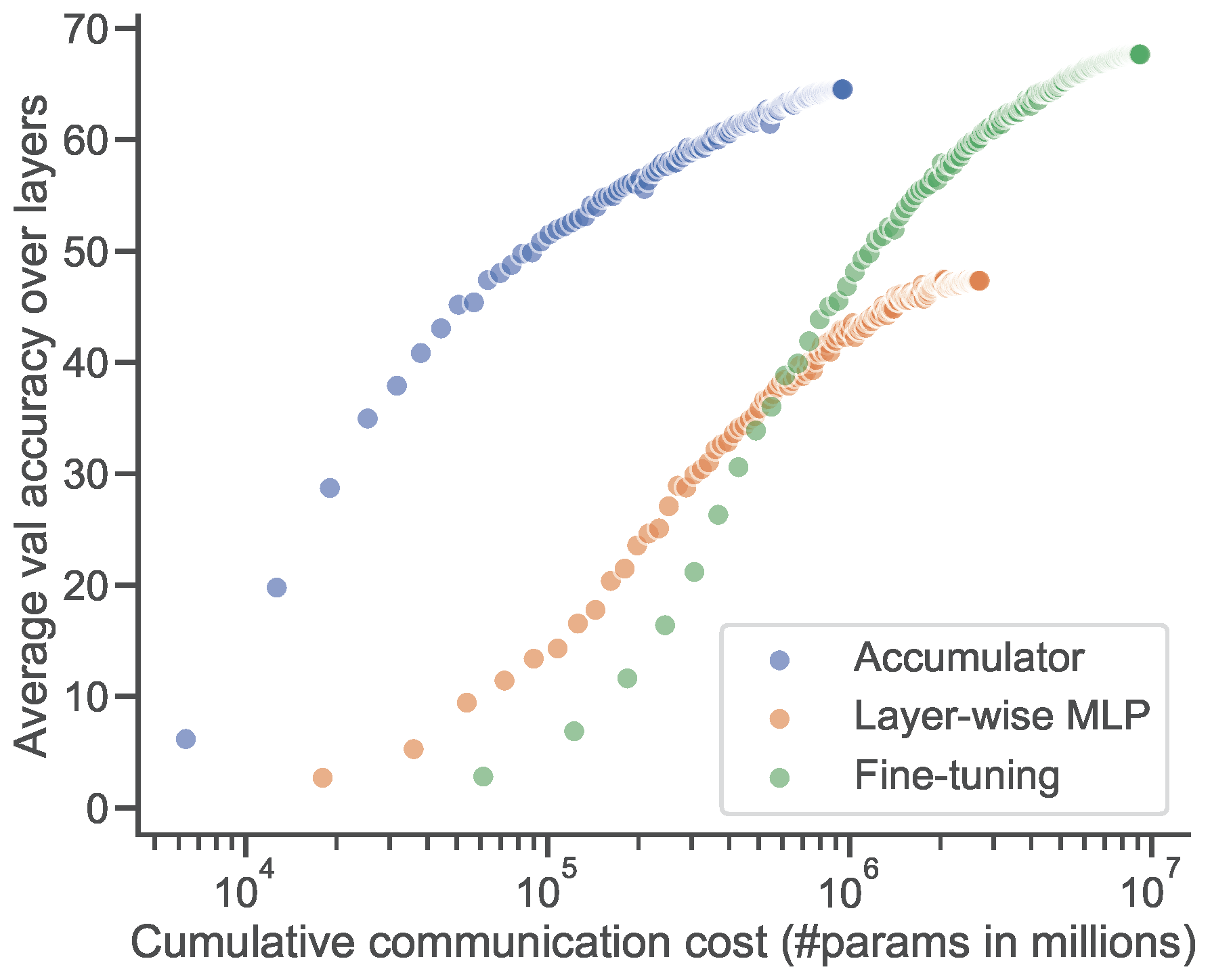}
        \caption{Near-IID case: $\alpha=1000$}
    \end{subfigure}
    \vspace{-5pt}
    \caption{{The effectiveness of \alg{}-based federated FM adaptation under anytime and anywhere setting in terms of communication cost and classification performance.} The experiments are conducted on CIFAR-100. Full details can be found in Section~\ref{sec:exp}. Each point corresponds to an evaluation during FL, where the cumulative communication cost measures the communication of gradients w.r.t. model parameters between clients and the server. We would like to emphasize that our \alg{} a) converges faster (less communication cost) than the baselines regardless of the data heterogenity condition and b) performs as well as the upper bound -- full-model fine-tuning.} 
    \label{fig:teaser}
\end{figure}

Given a pre-trained Transformer model, e.g. ViT~\cite{dosovitskiy2020image}, DeiT~\cite{Touvron21deit} and AST~\cite{gong21b_interspeech}, we re-wire its feature extraction pathway to tackle the anytime and anywhere challenges. Specifically, we keep track of the \texttt{CLS} tokens after each self-attention transformation and make use of the history of previous \texttt{CLS} tokens to revise the current \texttt{CLS} token by a lightweight Transformer, which is termed \emph{\alg{}}.   
%we replace its original classifier with our new task classifier and inject a novel attention-based adapter, termed \alg{}, into each block of the pre-trained Transformer. Our \alg{} modulates the outputs of all blocks of the pre-trained Transformer and provides a shortcut to the final classifier to make early predictions. Thus our system perfectly supports the FL training among clients of different compute capabilities. 
Our \alg{} has an order of magnitude fewer parameters than a pre-trained Transformer model and is the only module trainable during the local forward and backward propagations; therefore, the training and communication efficiencies can be both significantly improved. To show this, we make a comparison between \alg{} and a standard early-exit \citep{laskaridis2020hapi} model (\emph{Layer-wise MLP}, by inserting for each self-attention block a MLP classification head) and the full-model fine-tuning \citep{qu2022rethinking,nguyen2022begin}. The comparisons for non-IID, and IID cases are presented in Figure~\ref{fig:teaser}, which clearly show that our method is more efficient in reaching a certain target performance in both cases. In addition, due to the efficient optimization enabled by our \alg{}, user personalization for a particular client can be conducted efficiently with even better performance than fine-tuning the whole model.

% The pre-trained FM plus our adapter can be deployed into client devices by a server. Only the attention-based adapter is trainable during the local model training, with the rest parameters fixed. Our adapter has orders of magnitude fewer parameters than a pre-trained Transformer model, which significantly reduces the communication cost. Also, due to the adaptation of pre-trained Transformer FMs, our framework effectively reduces the effect of data heterogeneity among clients as per the finding in~\cite{qu2022rethinking}. 
% % Our novel adapter modulates the forward pass and can make early predictions.
% Our attention-based adapter (termed \emph{Accumulator}) \steve{adapter and accumulator are used a bit liberally and mean different things in our context} can modulate the outputs of all blocks of the pre-trained Transformer and can make early predictions at block granularity. 
% Thus, the Accumulator enables our FL framework to learn collaboratively among devices with heterogeneous computational constraints and infer at \emph{any time} of a forward pass. Furthermore, as the local training is remarkably accelerated, our Accumulator can be quickly personalized to a novel environment.

The contributions of our work are the follows:
\begin{itemize}
    \item We take a different perspective to the existing FL literature and propose a parameter-efficient learning method to adapt the pre-trained Transformer FMs in FL scenarios.
    \item We propose a novel parameter-efficient adapter, which modulates all layers of a pre-trained Transformer FM and allows flexible early predictions (anytime inference).    \item Extensive experiments on standard FL benchmarks, including CIFAR-100, FEMNIST and SpeechCommandsv2, show our method can improve global accuracy, personalization and communication efficiency with excellent robustness to data and compute heterogeneities.
\end{itemize}

% \tim{We should have a teaser figure for the start that shows accuracy vs miliseconds for anytime inference for one of our models vs a baseline, and/or a scatter plot of accuracy vs comms cost for a variety of the models we consider for one suitable experiment.}
\section{Related Work}

% \steve{Currently, there seems to be a lot of intersection between related work and intro. We can compress. Maybe in the introduction we can focus on why the problems of heterogneity are relevant in cross device and their impacts and then in related work we can define what has been done to mitigate them.}

\subsection{Federated Learning}
Though federated learning is still an emerging topic, there are already many works published in the literature. There exist two general FL settings, \emph{cross-device}~\cite{mcmahan2017communication} and \emph{cross-silo}~\cite{heikkila2020differentially} FL.
% \steve{Maybe briefly define them?}. 
In this paper, we focus on the former. The main research focuses in this setting are designing systems to solve communication efficiency, data and system heterogeneity
% \footnote{TODO:discuss privacy? PerFL?} 
problems. Researchers have proposed different techniques to improve communication efficiency. E.g. \cite{alistarh2017qsgd, fu2020don} proposed to use the quantized model parameters or gradients during the FL communication. Similarly, \cite{stich2018sparsified} sparsified, and \cite{yu2021adaptive,jiang2022model} pruned the training model into smaller cardinality to reduce communication cost. 
% \steve{Pruning can also affect compute/memory}. 
And some researchers tried to address this by progressive model training~\cite{wang2022progfed}. Another main focus is on data heterogeneity. In contrast to centralized training where the learner has access to the whole distribution, each worker having access to a biased distribution in non-i.i.d. FL negatively affects convergence and final model accuracy. In attempts to alleviate this, people proposed to add proximal regularization in the local training termed FedProx~\cite{li2020federated}. Alternatively, some normalized averaging technique to mitigate the inconsistency between different clients is proposed in~\cite{wang2020tackling}. Interestingly, more recently, researchers found that model initialization (pre-trained v.s. random) plays an important role in reducing the detrimental impact of  heterogenienty~\cite{nguyen2022begin}, and so does model architecture (Transfomer v.s. CNN)~\cite{qu2022rethinking}. System heterogeneity is also a concern in cross-device FL, where different participants may have different hardware resources and thus be unable to perform the same amount of learning. Some researchers used nested-dropout~\cite{rippel2014learning} as a nicely fit method to address the varied of computational constraints between different clients~\cite{horvath2021fjord}, as a different dimension from considering early-exit networks~\cite{laskaridis2020hapi}. 

Our work differs from the existing literature. Rather than training from scratch, we focus on the challenges and opportunities of adapting a pre-trained Transfomer FM by FL. While this might seem to exacerbate communication bottlenecks and system heterogeneity issues above, we show that with our novel adapter module we can simultaneously ameliorate all the above challenges of communication cost, heterogeneous device capabilities, and difficulty of federated learning on non-i.i.d. data. Our work goes substantially beyond  \cite{qu2022rethinking} and \citep{nguyen2022begin}, who just focus on conventional federated fine-tuning (which we treat as a baseline), to support communication efficient federated fine-tuning with support for heterogeneous device capabilities -- thanks to our novel adaptation module.

\subsection{Parameter-Efficient Learning}
The idea of parameter-efficient learning of adapters was first proposed in~\cite{rebuffi2017learning} for adapting a single model into multiple datasets. It has since been extended into various problems, including few-shot learning~\cite{li2022cross} and ASR~\cite{tomanek2021residual}. Especially
as the pre-trained large-scale FMs' model sizes are soaring significantly, parameter-efficient (transfer) learning has become important in NLP. Instead of fine-tuning the full pre-trained model, people lean toward designing different small set modules for adapting the pre-trained FMs into downstream tasks~\cite{li2021prefix, lester2021power, houlsby2019parameter}. \cite{li2021prefix} found that tuning the prompt input of the pre-trained language model enables excellent performance on downstream tasks. While \cite{houlsby2019parameter} found that fine-tuning some injected adapters can be more effective than fine-tuning the top layers of a pre-trained NLP model. In this work, we make the first attempt at parameter-efficient adaptation in an FL context, developing a novel transformer-based adaptation module specifically customized for this task. 

%as the first attempt, we cast federated learning as a downstream task of adapting a pre-trained Transfomer FM via a novel adapter.
% Unlike these works, our work focuses on designing a novel adapter module to efficiently adapt a pre-trained Transfomer FM into downstream federated learning.
\section{Preliminaries}
\subsection{Federated Learning\label{sec:fl}} 
Let us consider a typical setting of FL with $K$ devices,  
% \steve{Current setup defines cross-silo setting (no partial participation)}
where a local device $i$ has $N_i$ private training examples denoted by $\{(\mathbf{x}_j, y_j)\}_{j=1}^{N_i}$ with $\mathbf{x}_j$ the input image and $y_j$ the target label. The learning objective following \cite{mcmahan2017communication} aims at finding a model parameter $w$ that minimizes the weighted average loss over all local devices: 
% \steve{Point to FedAvg.}
\begin{equation}\label{eq:global}
    w = \underset{w}{\arg\min} \sum_{i=1}^K \alpha_i \mathcal{L}_i(w), \quad \mathcal{L}_i(w)= \frac{1}{N_i}\sum_{j=1}^{N_i}\ell(F_{w}(\mathbf{x}_j), y_j),
\end{equation}
where $\alpha_i=N_i/\sum_{i=1}^K N_i$, $\ell(F_{w}(\mathbf{x}_j), y_j)$ is the task-specific loss function and $F_{w}()$ is the formed model. %$(\mathbf{x}, y)$ are input and ground truth data pairs stored in each device locally. 
The main difference that renders the problem difficult is we can only compute $\mathcal{L}_i(w)$ on device $i$ to protect the privacy for the user. The common setup (see more details in \cite{beutel2020flower}) introduces a server to receive gradients sent from each client device, and therefore brings two major challenges: communication cost and data heterogeneity. 

\paragraph{Personalization.} \textcolor{black}{
Minimizing Eq.~\ref{eq:global} explicitly optimizes the generalization of the shared global model $w^*$ across all clients. However, the model performance on individual clients might still be sub-optimal, especially under client data heterogeneity.
%due to the limit of model capacity and randomness in stochastic gradient descent. 
Clients often care more about personalized performance (i.e., overfitting to client data). Therefore, given the outcome of global federated model learning in Eq.~\ref{eq:global} denoted by $w^*$, each client $i$ can then further fine-tune the parameters locally to obtain personalized parameter $w_i$
\begin{equation}\label{eq:per}
    w_i = \underset{w^*}{\arg\min}  \frac{1}{N_i}\sum_{j=1}^{N_i}\ell(F_{w^*}(\mathbf{x}_j), y_j),
\end{equation}
%where $w_i$ is the model personalized for user $i$.
}

\subsection{Parameter-Efficient Learning}
Parameter-efficient learning is a typical strategy for adapting  pre-trained FMs to downstream tasks. The full model size of an FM is often much larger than the size of downstream task data making the fine-tuning prone to overfitting. Not mentioning the back-propagation over the full FM is extremely expensive. 
%, and thus, fine-tuning the whole model for a downstream task is both computationally demanding, and easy to overfit on comparatively small downstream task data. Therefore, researchers designed 
To this end, multiple PE learning methods~\cite{houlsby2019parameter,rebuffi2017learning,tomanek2021residual,li2021prefix,lester2021power,zaken2021bitfit} have been proposed for fast adaptation of FMs, in which the learning objective is typically formulated as 
\begin{equation}\label{eq:pe}
    w_{\text{PE}} = \underset{w_{\text{PE}}}{\arg\min}  ~ \frac{1}{M}\sum_{j=1}^{M}\ell( F_{w_{\text{FM}}, w_{\text{PE}}}(\mathbf{x}_j), y_j),
\end{equation}
where $w_{\text{FM}}$ corresponds to the frozen pre-trained foundation model, $w_{\text{PE}}$ is the weight associated to the introduced
parameter-efficient module and \textcolor{black}{$\{(\mathbf{x}, y)\}^{M}$ are the $M$ data pairs from the downstream task of interest.} 
% $A$ here represents a operation which combines $w_{\text{FM}}$ and $w_{\text{PE}})$ a single model.
Our goal in this paper is to design such a lightweight module for Transformer-based FMs in the context of FL. 

\section{Adapting Transformer FMs into Federated Learning}
\begin{figure}[t]
     \centering
    \includegraphics[width=0.9\linewidth]{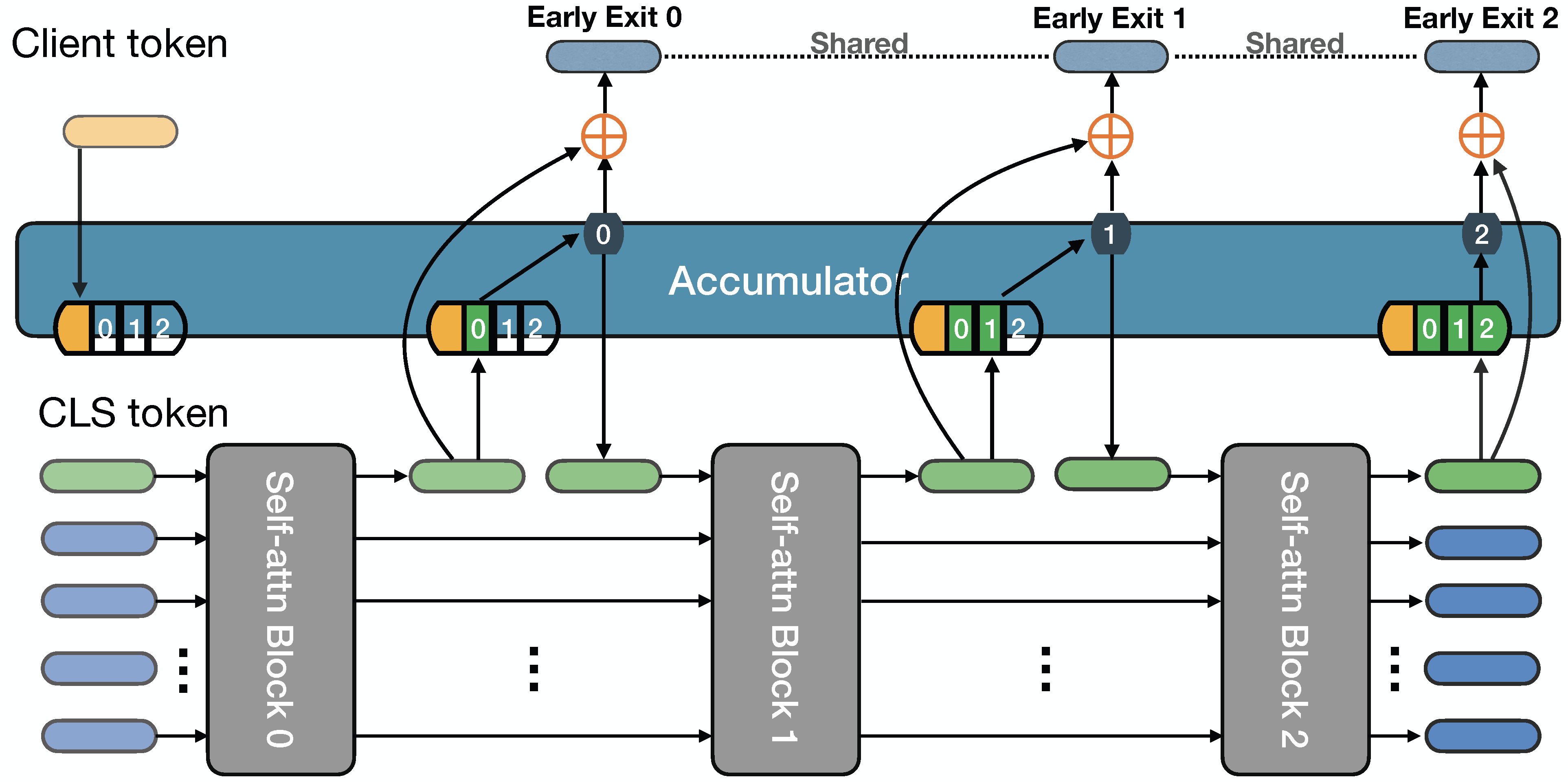}
    \vspace{-5pt}
    \caption{\textcolor{black}{Model architecture of the proposed \alg{}}. Given a frozen Transformer FM (in grey), the \alg{} aggregates the information from the history of the \texttt{CLS} tokens through the transformation, which yields a better feature representation for early exits at different layers. The early exit slots are numbered (there are three exits in this example), and the corresponding pathways are denoted by different line types (e.g., dotted for exit 0 and dashed for exit 1). A residual connection is introduced to each early exit between the \texttt{CLS} token at the exit and the output of the \alg{}. A special \texttt{CLS} token (or referred to as \texttt{Client} token) is introduced to be the first token of the \alg{}, which can be tuned for personalization while freezing the other parts.   
    }
    \label{fig:arch}
\end{figure}

The motivation of our work is to treat federated learning as a downstream task for adapting a pre-trained Transformer FM. In the following sections, we will introduce the two main modules, including a pre-trained Transformer and our attention-based adapter \alg{}, and how \alg{} modulates the outputs of a Transformer to support the early predictions. The overview of our model architecture is depicted in Figure~\ref{fig:arch}.

% \steve{TODO: Add a figure that describes the architecture and training workflow}

\subsection{Transformer model}
%This section will introduce how a pre-trained Transformer encoder can be used in our federated training.
A Transformer model typically consists of a sequence of residual blocks of multi-head self-attention (MSA), each followed by a residual block of feed-forward multilayer perceptron (MLP) with LayerNorm (LN) applied to both MSA and MLP blocks. 
Denote by $\mathbf{x}$ the input, $\mathbf{p}$ the positional encoding, and $\mathbf{z}^l := [z_{\text{cls}}^l, z_1^l, \ldots, z_N^l]$ the intermediate tokens,
the feed-forward pass of a Transformer can be formalized as
% \begin{equation}
\begin{align}
        \mathbf{z}^0 &= \text{Tokenizer}(\mathbf{x}) + \mathbf{p}, \label{eq:z0}\\
        \mathbf{z}^l &= \text{MSA}(\text{LN}(\mathbf{z}^{l-1})) + \mathbf{z}^{l-1}, \quad l=1\cdots L,  \label{eq:msa}\\
        \mathbf{z}^l &= \text{MLP}(\text{LN}(\mathbf{z}^{l})) + \mathbf{z}^{l}, \quad l=1\cdots L. \label{eq:mlp}
\end{align}
% \end{equation}
%where $\mathrm{x}_{1:N}$ is the tokenized input sequence with an additional class ([CLS]) token  $\mathrm{x}_{\text{cls}}$, $\mathbf{p}$ is the positional embedding added to the token sequence. $\mathbf{z}_l$ is the latent representation in layer $l$, in which $\mathbf{z}_l^0$ represents the transformed [CLS] token. After encoding the input sequence through all $L$ layers, typically, only the $\mathbf{z}_L^0$ is used for label inference in the classification tasks. 

%In summary, a Transformer encoder is a mapping function formed as
%\begin{equation}
%\begin{aligned}
%    \mathbf{z}_L = S^L_{\theta} \circ T(\mathbf{x}) = s^L_{\theta}\circ\cdots\circ s^1_{\theta} \circ T(\mathbf{x}),
    % \text{where} ~~~ F_{\theta} (\mathbf{z}_0) &= F^L_{\theta}\circ\cdots\circ F^1_{\theta} (\mathbf{z}_0), \\
    % F^l_{\theta} (\mathbf{z}_0) &= \text{MLP}^*(\text{LN}(\text{MSA}^*(\text{LN}(\mathbf{z}_{l-1}))
%\end{aligned}
%\end{equation}
%where $s^l_{\theta} (\mathbf{z}_{l-1})$ is computed as alternating Eq.~\ref{eq:msa}, \ref{eq:mlp} and $\theta$ is the model parameters and $T(.)$ is the tokenization function to obtain $\mathbf{z}_0=T(\mathbf{x})$.

% \steve{Define somewhere the adapter.}

\subsection{\alg{}}
\label{sec:accumulator}
To adapt a pre-trained Transformer model, 
we inject our Accumulator into each\footnote{Note that there is only one Accumulator that handles the outputs from all layers.} self-attention block followed by a shared MLP head to enable early predictions. Formally, by collecting the history of \texttt{CLS} tokens, we replace $z_{\text{cls}}^l$ by 
\begin{align}
    z_{\text{cls}}^l \leftarrow h^l_{l+1} \quad \text{with} \quad h^l = \text{\alg{}}_{\phi}([z_{\text{client}}, z_{\text{cls}}^0 + p'_{0}, \ldots, z_{\text{cls}}^l + p'_{l}]),
\end{align}
where the \alg{} parameterized by $\phi$ is another Transformer (randomly initialized) with a single\footnote{We find in Table~\ref{tab:paraTable1} that increases from a single block to 3 blocks yields little improvement in accuracy but a significantly heavier communication burden.} self-attention block as defined by Eq.~\ref{eq:msa} and ~\ref{eq:mlp}. $h^l_{l+1}$ means the $l+1$ element in $h^l$. In our \alg{}, the positional embedding $\mathbf{p}' := [p'_{0}, \ldots, p'_{L}]$ embeds the layer information from a pretrained Transformer. Note that the modified $z_{\text{cls}}^l$ will be again used by the pre-trained Transformer, specifically, the next self-attention block, to produce $\mathbf{z}^{l+1}$.

For early predictions, instead of taking the modified $z_{\text{cls}}^l$, we find it is better to make use of the client token and the original $z_{\text{cls}}^l$, specifically, the task-specific prediction at exit $l$ is given by 
\begin{align}
   \hat{y}^l =  \text{MLP-head}_{\psi}(h^l_0 + z_{\text{cls}}^l),
\end{align}
where MLP-head parameterized by $\psi$ is the early exits shared across all layers.

\subsection{Federated learning of \alg{}}
Following the PE optimization in Eq~\ref{eq:pe}, the learnable parameters $w_{\text{PE}} = (\phi, \psi)$ reduces to the weights of the \alg{} and the MLP head. 

\paragraph{Learning.}
Now the learning problem in Eq.~\ref{eq:pe} becomes 
\begin{equation}\label{eq:flacm}
    \underset{w_{\text{PE}}}{\arg\min}  \sum_{i=1}^K  \frac{\alpha_i}{M}\sum_{j=1}^{M} \ell(\hat{y}^{l_i}_j, y_j),
\end{equation}
where $l_i$ is the layer of an early exit for client $i$, which 
is set for each inference in client $i$. Assuming the client $i$ has the budget using up to $L_i$ layers, there are two schemes for choosing the value of $l_i$: a) $l_i = L_i$ and b) $l_i$ is taken uniformly at random from the range $\{1, \ldots, L_i\}$. 
We attribute this property of our formulation as \emph{anytime} to accommodate clients with different computing and memory capacities.

\paragraph{Personalization.} There are two choices: fine-tuning the whole \alg{} $(\phi, \psi)$ or the \texttt{Client} token $z_{\text{client}}$ only. Given that they are both lightweight, the personalization is less likely to overfit even under an extremely low data regime.

\paragraph{Inference.} Given an FL trained model in client $i$, it can infer labels $\hat{y}^{L_i}$ not only from block $L_i$ according to its capability but also labels $\hat{y}^{l_i}$ as long as $l_i$ is smaller than the budget constraint $L_i$. This property can be very useful in the case when a cellphone suffers from low battery, which enables the user to set a small budget $l_i<L_i$ to allow energy-efficient inference.

% Interestingly, as our $G_\omega \circ F^M_{\psi}(t(F^{\color{red}{L_i}}_{\theta}(\mathbf{x})))$ can be trained flexibly with different $L_i$ in FL. At inference, the user can set a $L_i' <= L_i$ whenever they want. For example, when the user device, such as a cellphone, is used with a (low) battery, they can set small $L_i'$ for fast inference to save battery cost. And importantly, since our \alg{} $G_\omega \circ F^M_{\psi}$ is parameter efficient, it can be fast personalized following Eq.~\ref{eq:per} in a local device after the FL training.

\section{Experiments} \label{sec:exp}

\subsection{Experimental Setup}

\paragraph{Dataset Settings and Implementation.} To verify the efficacy of our proposed FL framework, we conduct experiments on the standard FL benchmarks with Flower codebase~\cite{beutel2020flower}, including CIFAR-100~\cite{cifar}, FEMNIST~\cite{emnist} and SpeechCommandV2~\cite{speechcommands}, as downstream tasks of pretrained DeiT~\cite{Touvron21deit} and AST~\cite{gong21b_interspeech}. 
We use the original test set in CIFAR-100 as the global test dataset and split 10,000 images from the original training set as the personal validation set for each client when conducting the personalization experiments on CIFAR-100-C.
We simulate three different data partitions for CIFAR-100, including one IID-data partition ($\alpha=1000.0$), and two non-IID data partitions($\alpha=1.0, 0.1$) by LDA partitioning following prior works~\cite{karimireddy2020scaffold,wang2020federated} with 100 clients. FEMNIST has a total of 80,000 images (grayscale) of 62 different character classes (10 numeric, 26 lowercase, and 26 uppercase).
For FEMNIST, we partition the training data in two ways, one IID and one non-IID, as in~\cite{emnist} with 381 clients. The Speech Commands v2 dataset consists of 105,829, 16KHz 1-second long audio clips of a spoken word. And we conduct our experiment with the 12-classes version, with 10 classes as "yes", "no",
"up", "down", "left", "right", "on", "off", "stop", "go" and one class "unknown" and a class "silence" which has no spoken word in the clip. The dataset has
three disjoint sets of speakers: 2112 for training, 256 for validation and 250 for the test.

For each round in FL, we sample $10\%$ of clients for training on CIFAR100 and FEMNIST, and $1\%$ on SpeechCmdv2.
We use a pretrained DEiT-small~\cite{Touvron21deit} model with $16\times16$ patches on ILSVRC-2012~\cite{Fei-Fei2010} as the foundation model, which can be used for both image and speech recognition -- i.e. the backbone of AST~\cite{gong21b_interspeech}.

All experiments are conducted with Pytorch on a single Nvidia Tesla V100 GPU with results repeated three times and reporting the mean and std.
All models are trained using an SGD optimizer with a cosine annealing learning rate schedule. {The detailed recipe can be found in the Appendix~\ref{app:settings}.}

\paragraph{Early Exits.}
Each client has its own computing capability. According to their capability, they can support a certain amount of FLOPs for one inference, or one FL round. For example, given a pretrained Transformer, some clients may be able to pass the input data through all layers of the Transformer to train a classifier. Some could only pass through half of them to the same end. 
Specifically, how many layers of a pre-trained transformer can be passed through during the local model training for a client depends on the client's computing capability. One can train a classifier head to make predictions for each layer to support the different compute capabilities among clients. These layer-wise classifier heads are often called early exits~\cite{laskaridis2021adaptive, leontiadis2021s}. For example, we can straightforwardly train 12 early exit classifiers for a given pretrained DeiT-small, which has 12 layers. And then, those exits can be used for different clients accordingly. Briefly, we can treat 12 exits corresponding to  12-tiers of client compute capability -- namely, the different levels of computing budgets in terms of anytime inference. 

% In this way, we simply treat how many layers a client can pass through as the `virtual' tier value of the compute capability of a client device~\cite{horvath2021fjord}, e.g. there are 12 tiers when we use a DeiT-small model, which has 12 layers.

% 1) the conventional FL training and test, where clients train the local models by passing through all layers of a pretrained transformer.

% with the same tier of maximum compute capability. During the FL training, each client infers through the full training model using the maximum capability; 2) FL training for anytime inference, where an FL model is trained on clients with the same tier of maximum compute capability. During the training, each client infers through the training model partially using a random level of partial capability. And at the test stage, ; 3)

% In federated learning, the connected clients can have different local data distributions and heterogenous compute capabilities. Thus, it requires that the FL system should be robust to the data heterogeneity across clients. 

% Our progressive experiments evaluate Accumulator in three aspects: 
% 1) Adaptability to the foundation model;
% 2) Inference Performance at Anytime;
% 3) Anywhere personalized adaptation.

\begin{table}[t]
\caption{Conventional FL performance.}
\centering
\scalebox{0.8}{
\begin{tabular}{l|ccc|cc|c}
\toprule
\multirow{2}{*}{Method}\Tstrut & \multicolumn{3}{c|}{CIFAR-100} & \multicolumn{2}{c|}{FEMNIST} & \multirow{2}{*}{SpeechCmdV2} \\ \cmidrule{2-6}
& IID ($1000.0$) & Non-IID ($1.0$) & Non-IID ($0.1$) & IID & Non-IID  \\ \midrule
Fine-tuning & \gray{85.05$\pm$0.55} & \gray{84.91$\pm$0.98} & \gray{84.25$\pm$1.14} & \gray{86.87$\pm$0.74} & \gray{85.28$\pm$1.45} & \gray{98.22} \\
\midrule
Linear head & 74.37$\pm$0.42 & 73.85$\pm$0.74 & 72.79$\pm$0.83 & 74.02$\pm$1.33 & 72.35$\pm$0.52 &69.47\\
+ {PA}   & 83.28$\pm$0.67 & 82.57$\pm$1.12 & 81.41$\pm$1.28  & 82.34$\pm$1.56 & 81.60$\pm$0.66 &  94.74\\
\midrule
MLP head    & 76.49$\pm$0.43 & 75.69$\pm$0.81 & 74.44$\pm$0.95  & 74.26$\pm$1.69& 72.87$\pm$0.69 & 74.33\\
+ {PA}   & 84.55$\pm$0.69 & 83.87$\pm$1.25 & 82.33$\pm$1.24  & 82.49$\pm$1.88 & 81.54$\pm$0.78 & \textbf{95.63}\\
\midrule
\alg{}      & 84.90$\pm$0.75 & 84.34$\pm$1.33 & 83.31$\pm$1.54  & 79.12$\pm$1.64 & 78.05$\pm$0.75 & 93.24\\
+ {PA}   & \textbf{85.35}$\pm$0.49 & \textbf{85.11}$\pm$1.28 & \textbf{84.02}$\pm$1.32 & \textbf{83.68}$\pm$1.75 & \textbf{82.24}$\pm$0.84 & \textbf{95.27}\\
\bottomrule
\end{tabular}
}
\label{tab:exp1}
\end{table}
\begin{table}[t]
% \vspace{-2em}
\caption{Anytime FL performance. 
% \steve{Budgeted and anytime mean different things. See MSDNet paper ...}
}
\centering
\scalebox{0.8}{
\begin{tabular}{l|ccc|cc|c}
\toprule
\multirow{2}{*}{Method}\Tstrut & \multicolumn{3}{c|}{CIFAR100} & \multicolumn{2}{c|}{FEMNIST} & \multirow{2}{*}{SpeechCmdV2}\\ \cmidrule{2-6}
& IID ($1000.0$) & Non-IID ($1.0$) & Non-IID ($0.1$) & IID & Non-IID \\ \midrule
Fine-tuning & \gray{67.24$\pm$1.13} & \gray{66.95$\pm$1.93} & \gray{60.05$\pm$2.51} & \gray{76.43$\pm$1.45} & \gray{75.82$\pm$2.42} & \gray{93.71} \\
\midrule
L.W. Linear & 36.42$\pm$0.80 & 36.06$\pm$1.35 & 34.00$\pm$1.77 & 35.92$\pm$1.23 & 35.49$\pm$1.74 & 65.10 \\
+ {PA} & 47.33$\pm$0.95 & 47.94$\pm$1.53 & 47.11$\pm$1.92 & 57.36$\pm$1.45 & 56.78$\pm$1.89 & 84.16\\
\midrule
L.W. MLP & 37.62$\pm$0.76 & 37.84$\pm$0.98 & 38.22$\pm$1.34 & 35.27$\pm$1.29 & 34.54$\pm$1.78 & 64.57 \\
+ {PA} & 48.65$\pm$0.89 & 48.39$\pm$1.02 & 47.94$\pm$1.21 & 55.21$\pm$1.48 & 54.42$\pm$2.01 & 83.19\\
\midrule
\alg{} & 64.28$\pm$1.01 & 63.02$\pm$1.79 & 57.34$\pm$2.23 & 75.47$\pm$1.41 & 75.12$\pm$1.86 & 84.66\\
+ {PA} & \textbf{65.23}$\pm$1.26 & \textbf{64.28}$\pm$1.84 & \textbf{58.40}$\pm$1.67 & \textbf{76.55}$\pm$1.63 & \textbf{76.03}$\pm$2.13 & \textbf{88.30}\\
\bottomrule
\end{tabular}
}
\label{tab:exp2}
\end{table}

\paragraph{Baselines.} We run all our FL experiments with the  FedAvg~\cite{mcmahan2017communication} strategy. We compare with all the baselines, namely,
\textit{i})~\textit{Fine-tuning}~\cite{qu2022rethinking}, where the whole pretrained vision transformer is finetuned end-to-end in FL;
\textit{ii})~\textit{Layer-wise Linear} (L.W. Linear), where we append a learnable linear head after each transformer block\footnote{We also tried with shared linear head among layers, but it achieved worse results.} to be learned by FL, while keeping the rest frozen;
\textit{iii})~\textit{Layer-wise MLP} (L.W. MLP), where we append a two-layer MLP with GELU and Random Dropout after each transformer block, with the rest frozen. 
\textit{iv})~Our \textit{\alg{}}, where we append one shared \alg{} and one shared MLP into each transformer block, with the rest frozen. 
We also compare the variants of L.W. Linear, MLP, and \alg{} with {Parallel Adapters (PA), as proposed in~\cite{he2021towards}}, injected into the feed-forward MLP networks of the pretrained Transformer as their efficient and effective adaptation~\cite{he2021towards}.

\paragraph{Evaluation Settings.}
We evaluate all the methods in four different settings, including (1) \emph{Conventional FL} -- Clients train the local models by using only the final exit of a pretrained transformer, and the trained global model will be evaluated at a test set using only the final exit.
(2) \emph{Anytime FL} -- Clients train the local models by using a random exit at each iteration, and the trained global model will be tested at each exit.
(3) \emph{multi-tier FL} -- Clients train the local models by using a specific early exit determined by the tier of each client. The trained global model will be tested at each exit,
and (4) Personalization, where the FL-trained model from setting (3) will be further finetuned using the local data in each client. 

% The adapters can be employed in cooperation with these methods to better adapt the FMs parameter-efficiently.

\subsection{Experimental Results}

\paragraph{Conventional Federated Learning.}
Results in Table~\ref{tab:exp1} show the comparison between all competitors. The results show that fine-tuning works the best among all methods. This is unsurprising, as it has access to the large combined dataset of all clients and can use this to tune the whole model, but it incurs the most local training and communication costs. Among the other competitors, we can see that our \alg{} achieves the best results already when used alone. Parallel adapters are quite effective in enabling federated adaptation of the pretrained Transfomer, boosting the performance of base methods L.W. Linear and MLP significantly in all settings, especially on SpeechCmdv2. Nevertheless, our \alg{} is complementary with parallel adapters, achieving the best result in all cases, with some results even surpassing the Fine-tuning method, such as on CIFAR-100 with IID and Non-IID (1.0). The results demonstrate the efficacy of our proposed \alg{} to adapt a pretrained Transformer model into FL at anywhere under any type of data heterogeneity.

% The ability of efficient adaptation to full foundation models (FMs) is often required because the designed modules should fully use FMs.
% In addition, the adaption performance on full FMs is closely related to downstream and personal tasks.
% Therefore, we take the full fine-tuning as the base performance often used as a target in PE methods.
% Similarly, the "adaptation accuracy" is calculated in two steps:
% (1) froze all paramters in FMs;
% (2) tuning the added trainable parameters.
% In this setting, the total number of training rounds is 500, we only take the last layer as the exit.
% Table~\ref{tab:exp1} describes the adaptation accuracy according to the different data heterogeneityo
% Compared to linear and MLP heads, Accumulator achieves about the same accuracy as fine-tuning without inserting any parameters in FM, which is the expected performance of reusing the previous class tokens.
% If we use both the accumulator and adapter, the performance will be improved consistently and even surpass fine-tuning, \textit{i.e.,} such as $\alpha=1.0$ and $\alpha=1000.0$.
% In a word, efficient use of all class tokens of FMs will substantially improve the adaptation effect of PE fine-tuning in the conventional classification task.

\begin{figure*}[t]
    \centering
    \includegraphics[width=0.9\linewidth]{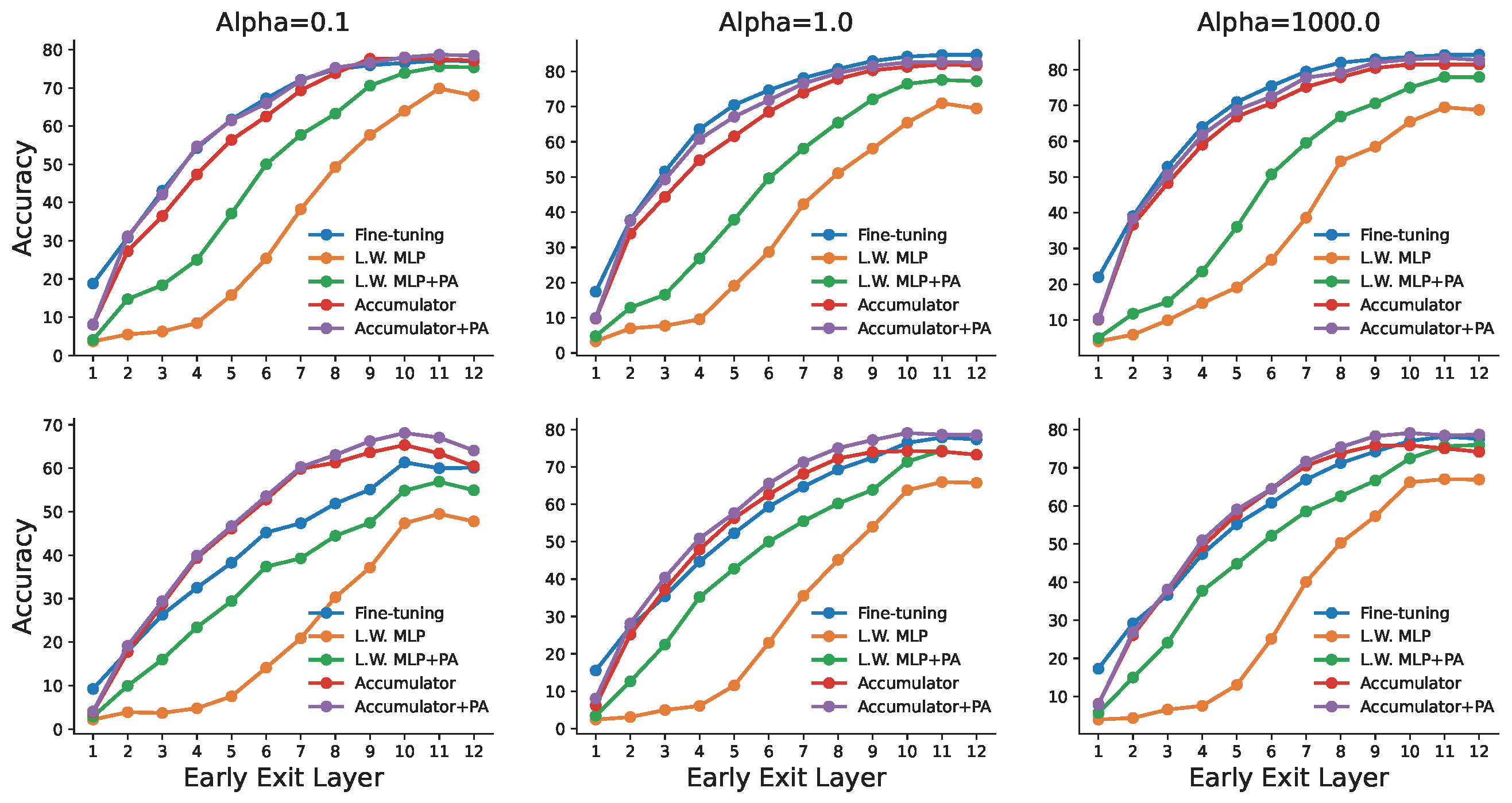}
    \vspace{-5pt}
    \caption{
        Test accuracy of each exit trained on CIFAR-100 with anytime federated learning (Top) and multi-tier constraint (Bottom).
        {The memory and FLOPs cost of each exit is given in Table~\ref{tab:budget}.}
        % \red{The total computation budgets (FLOPS) is 4.52G, corresponding to the first 11 exits: 0.42G, 0.78G, 1.15G, 1.51G, 1.88G, 2.26G, 2.63G, 3.01G, 3.38G, 3.76G, 4.14G.}
    }
    \label{fig:multi_tier}
\end{figure*}

\paragraph{Anytime Federated Learning.}
In this setting, we will consider the early exit situations. For each layer of a pretrained DeiT, we train an early exit, such as a linear head, an MLP head and our \alg{}. At each local training iteration, an exit layer index $l \in [0,\cdots,11]$ will be sampled randomly, and only that exit (and accumulator where relevant) will be trained. For fine-tuning, we append the layer-wise MLP heads and train all parameters during the FL training. After training, the global model will be evaluated on the test set at all exits. The results in Table~\ref{tab:exp2} and Figure~\ref{fig:multi_tier} \textit{top} show the average and budget-wise performance over all exits. Without surprise, we can see that global fine-tuning achieves the best at substantial comms cost. Among the more efficient competitors, we can see that our \alg{} outperforms L.W. Linear and MLP significantly when no extra adapters are used. With parallel adapters, all methods enjoy a performance boost, leading our \alg{} to the best performance in all cases, outperforming fine-tuning in a few situations.

% According to the architecture of ViTs, each layer is regarded as an exit. 
% In the training stage, to give the model the ability of anytime prediction, we sample one layer randomly at each client iteration as the current exit for the downstream task.
% The total number of rounds is set to 1500 and for each iteration on clients, an exit layer is sampled randomly in $[0, 11]$.
% As shown in Table~\ref{tab:exp2}, we compare the mean accuracy across all exit layers where the fine-tuning is colored grey because of the size of trainable parameters.
% The accumulator achieves the best anytime prediction performance close to finetuning at a smaller number of parametric levels.

\paragraph{Multi-tier based Federated Learning.} More realistically, there is a certain level of system heterogeneity among client devices. Individual devices in FL training  have a certain level of computing capability and their associated early exit should be persistently fixed in both training and testing. Results in the setting are evaluated and reported in Table~\ref{tab:exp3}, where we can see  that most results dropped to some extent compared with Table~\ref{tab:exp2}. This is expected due to the existence of system heterogeneity among clients. And most observations in Table~\ref{tab:exp3} are similar to the previous tables. One interesting observation is that in this harder scenario, our \alg{} outperforms the fine-tuning baseline consistently in all situations in CIFAR-100. Figure~\ref{fig:multi_tier}~\textit{bottom} shows the corresponding results for clients of different tiers on CIFAR-100 after training by different algorithms. The accumulator+adapter architecture performs most favourably. This again shows our \alg{} works for anytime inference in a harder and more practical scenario. {We also provide communication efficency in Appendix~\ref{app:exps}.}

% In a realistic scenario, clients may have different compute capabilities.
% Therefore, configuring a fixed exit layer for each client for inference in the federated training is more reasonable and practical.
% Under this restriction, we again measured the average accuracy of all exit layers of the previous method.
% Differently, we found that the accumulator with adapter outperformed fine-tuning overall, especially when $\alpha=0.1$, as shown in Table~\ref{tab:exp3}. 
% According to Figure~\ref{fig:multi_tier} we find that accumulator is ahead of fine-tuning by a relatively large margin in the middle of the budget.
% But we also have to admit that the first few budgets need more trainable parameters for better results.
% In other words, putting aside the enormous computational parameters of fine-tuning, which requires a large number of various clients with different computational capabilities to perform as it should, the accumulator can cope with this practical situation more comfortably.

\begin{table}[t]
\caption{Multi-tier based FL performance.}
\centering
\scalebox{0.8}{
\begin{tabular}{l|ccc|cc|c}
\toprule
\multirow{2}{*}{Method}\Tstrut & \multicolumn{3}{c}{CIFAR100} & \multicolumn{2}{|c|}{FEMNIST} & \multirow{2}{*}{SpeechCmdV2}\\ \cmidrule{2-6}
& IID ($1000.0$) & Non-IID ($1.0$) & Non-IID ($0.1$) & IID & Non-IID \\ \midrule
Fine-tuning       & \gray{58.17$\pm$0.52} & \gray{56.70$\pm$0.83} & \gray{42.47$\pm$1.41} & \gray{74.79$\pm$0.75} & \gray{74.36$\pm$1.36} & \gray{93.03} \\
\midrule
L.W. Linear & 32.43$\pm$0.34 & 30.98$\pm$0.60 & 21.33$\pm$0.97 & 35.68$\pm$0.54 & 35.17$\pm$1.01 & 65.14\\
+ {PA} & 46.96$\pm$0.41 & 45.14$\pm$0.68 & 33.94$\pm$1.03 & 57.06$\pm$0.60 & 54.67$\pm$1.15 & 83.23\\
\midrule
L.W. MLP & 36.57$\pm$0.32 & 34.38$\pm$0.58 & 24.37$\pm$0.93 & 33.12$\pm$0.59 & 32.88$\pm$1.13 & 64.84\\
+ {PA} & 51.33$\pm$0.47 & 49.29$\pm$0.66 & 36.29$\pm$1.07 & 51.88$\pm$0.67 & 51.39$\pm$1.27 & 82.28\\
\midrule
\alg{} & 58.27$\pm$0.56 & 57.32$\pm$0.88 & 47.26$\pm$1.18 & 72.14$\pm$0.72 & 71.33$\pm$1.39 & 85.21\\
+ {PA} & \textbf{59.33}$\pm$0.72 & \textbf{58.48}$\pm$0.96 & \textbf{48.55}$\pm$1.30 & \textbf{72.67}$\pm$0.81 & \textbf{72.05}$\pm$1.57 & \textbf{87.42}\\
\bottomrule
\end{tabular}
}
\label{tab:exp3}
\end{table}

\begin{table}[t]
\caption{Personalized performance on CIFAR-100-C and corrupted SpeechCmdv2 with multi-tier constraints after FL training on clean data (figures in bracket show relative improvement to performance before personalization).}
\centering
\resizebox{0.9\textwidth}{!}{
\begin{tabular}{l|ccc|c}
\toprule
Update part for\Tstrut & \multicolumn{3}{c|}{CIFAR-100-C} &\multicolumn{1}{c}{SpeechCmdV2} \\ 
\cmidrule{2-5}
  personalization & IID ($1000.0$) & Non-IID ($1.0$) & Non-IID ($0.1$) & Bkg. Noise (0.6) \\ \midrule
Full Model & \gray{34.30$\pm$1.85\tiny{(+32.19)}} & \gray{37.75$\pm$2.21\tiny{(+36.36)}} & \gray{26.94$\pm$3.14\tiny{(+25.85)}} & \gray{90.21}\tiny{(+8.59)} \\
\midrule
L.W. Linear +{PA} & 34.67$\pm$1.21\tiny{(+32.10)} & 37.55$\pm$1.79\tiny{(+35.21)} & 26.63$\pm$2.21\tiny{(+24.46)} & 80.82\tiny{(+6.40)}\\
{PA} Only & 34.06$\pm$1.01\tiny{(+31.49)} & 37.02$\pm$1.27\tiny{(+34.68)} & 25.96$\pm$1.58\tiny{(+23.79)}  &81.40\tiny{(+6.98)} \\
\midrule
L.W. MLP + {PA} & 31.69$\pm$1.42\tiny{(+29.46)} & 36.95$\pm$2.05\tiny{(+34.80)} & 22.30$\pm$2.96\tiny{(+20.26)} & 80.59\tiny{(+6.42)} \\
{PA} Only & 33.01$\pm$1.18\tiny{(+31.73)} & 37.38$\pm$1.34\tiny{(+35.23)} & 26.59$\pm$1.62\tiny{(+24.55)}  & 80.82\tiny{(+6.65)} \\
\midrule
\alg{} + {PA} & 38.23$\pm$1.72\tiny{(+34.78)} & 40.05$\pm$2.34\tiny{(+36.90)} & 31.26$\pm$2.98\tiny{(+28.30)}  &\textbf{85.46}\tiny{(+6.47)} \\
{PA} Only & 37.33$\pm$1.37\tiny{(+33.88)} & 39.45$\pm$1.48\tiny{(+36.30)} & 30.24$\pm$1.69\tiny{(+28.20)} & 85.40\tiny{(+6.41)} \\
\texttt{Client} Token Only & \textbf{45.38}$\pm$0.96\tiny{(+41.93)} & \textbf{47.02}$\pm$1.13\tiny{(+43.87)} & \textbf{38.25}$\pm$1.30\tiny{(+35.29)} & 85.24\tiny{(+6.25)} \\
\bottomrule
\end{tabular}
}
\label{tab:exp4}
\end{table}

% \begin{table}[t]
% \caption{Personalized performance on CIFAR-100-C and corrupted SpeechCmdV2 with multi-tier constraints after FL training on clean data.}
% \centering
% \resizebox{1.0\textwidth}{!}{
% \begin{tabular}{l|ccc|c|c}
% \toprule
% Update part for\Tstrut & \multicolumn{3}{c|}{CIFAR-100-C} &\multicolumn{1}{c|}{SpeechCmdV2} &\multirow{2}{*}{Parameter Size} \\ 
% \cmidrule{2-5}
%   personalization & IID ($1000.0$) & Non-IID ($1.0$) & Non-IID ($0.1$) & Bkg. Noise (0.6) & \\ \midrule
% Full Model & \gray{34.30$\pm$1.85} & \gray{37.75$\pm$2.21} & \gray{26.94$\pm$3.14} & \gray{90.21}  & 30.62M \\
% \midrule
% L.W. Linear + Adapter & 34.67$\pm$1.21 & 37.55$\pm$1.79 & 26.63$\pm$2.21 & 80.82& 1.06M \\
% Adapter Only & 34.06$\pm$1.01 & 37.02$\pm$1.27 & 25.96$\pm$1.58  &81.40& 0.60M \\
% \midrule
% L.W. MLP + Adapter & 31.69$\pm$1.42 & 36.95$\pm$2.05 & 22.30$\pm$2.96&80.59 & 9.55M \\
% Adapter Only & 33.01$\pm$1.18 & 37.38$\pm$1.34 & 26.59$\pm$1.62  & 80.82& 0.60M \\
% \midrule
% \alg{} + Adapter & 38.23$\pm$1.72 & 40.05$\pm$2.34 & 31.26$\pm$2.98  &\textbf{85.46}& 3.77M \\
% Adapter Only & 37.33$\pm$1.37 & 39.45$\pm$1.48 & 30.24$\pm$1.69 &85.40& 0.60M \\
% CLS Token Only & \textbf{45.38}$\pm$0.96 & \textbf{47.02}$\pm$1.13 & \textbf{38.25}$\pm$1.30 &85.24 & 0.38K \\
% \bottomrule
% \end{tabular}
% }
% \label{tab:exp4}
% \end{table}

\paragraph{Personalization.}
Since clients' class and data distributions are typically different and diverse, adapting the globally trained model through per-client personalization is an interesting task for FL.
Based on the pertained model of multi-tier federated learning as the most practical scenario, we can adapt the model to the local data of each client efficiently.
Specifically, the multi-tier FL pretrained model is fine-tuned for ten epochs for each client. We use noisy data for personalization and the final test to simulate personal data distributions. For the corrupted CIFAR-100, we simply use CIFAR-C~\citep{hendrycks2019cifarC}. To corrupt SpeechCmdv2 dataset, we add $60\%$ background noise into each validation speaker and split them into two sets for personalization and the final test. From the results in Table~\ref{tab:exp4}, we can see that when tested on the corrupted data, the performances of all methods degraded. Now, our \alg{} shows outstanding performance among all competitors, including updating the full model during personalization. More interestingly, updating the \texttt{Client} token, which has an extremely small set of parameters, i.e. 0.38K parameters according to Table~\ref{tab:paraTable1} (Left), in our \alg{}, works overall the best among all situations. And also, when comparing the test accuracy gains after personalization, we can see our \alg{} produces the most, especially when the \texttt{Client} token is solely used for personalization fine-tuning.

% to make a unique data processing on each client's local data, which shifts the data distribution between clients.
% As shown in Table~\ref{tab:exp4}, the performance gap is further enlarged.
% It is worth noting that the accumulator only requires tuning a client token of size 0.384K to achieve outstanding personalized performance that significantly exceeds full fine-tuning.

% \input{tabs/exp4.tex}

\begin{table}[t]
\caption{Left: parameter number of different methods. Right: Ablation study of \alg{}.}
\centering
% \scalebox{0.94}{
\begin{tabular}{l|c}
\toprule
\multirow{1}{*}{Method}\Tstrut & Parameter Num\\
\midrule
Full Fine-tuning & 30.62M\\
\midrule
{Parallel Adapter} & 0.60M \\
\midrule
Layer-wise Linear & 0.46M\\
\midrule
Layer-wise MLP & 8.95M\\
\midrule
\alg{} &  3.17M\\
\texttt{Client} Token & 0.38K \\
\bottomrule
\end{tabular}
% }
\label{tab:paraTable1}
% \scalebox{0.90}{
\begin{tabular}{l|c}
\toprule
\multirow{1}{*}{}\Tstrut & Mean performance\\
\midrule
Full Accumulator &  \multirow{1}{*}{48.55}\\
\midrule
\# self-attn blocks=3 & 48.86 (+0.31)\\
\midrule
No Replace & 45.81 (-2.74) \\
\midrule
No Residual & 43.34 (-5.21)\\
\midrule
No {Parallel Adapter} & 47.93 (-0.62)\\
\midrule
Linear Head & 44.6 (-3.95)\\
\bottomrule
\end{tabular}
% }
\end{table}

% \subsection{Further Analysis}

\paragraph{Ablation Study.}
In order to verify the contributions of different components proposed in our \alg{}, ablation studies were conducted based on the multi-tier based FL setup on CIFAR-100 ($\alpha=0.1$), the hardest scenario. 
In Table~\ref{tab:paraTable1} (right), the full accumulator indicates a transformer with depth=1, class token parallel connection, token replace mechanism, parallel adapter, and an MLP classification head.
Among them, we find that parallel connection influences performance the most. 
This is mainly because DeiT's original class tokens are well learned on ImageNet, which are directly beneficial for model predictions in the downstream tasks. Removing token replacing reduces the effect of modulating the extracted features from DeiT. Thus the performance drops noticeably. And we also tried larger depth, i.e. 3, in the Accumulator, but it seemed not to produce much performance gain. Our \alg{} works complementarily with parallel adapters used but apparently does not rely on them to have excellent performance.

% in the initial adaptation stage of the accumulator provide a significant benefit in prediction, which speeds up the accumulator's adaptation.
% The lack of Replace drastically reduces the number of interactions between the accumulator and FM, which directly affects the efficiency of the accumulator's use of class tokens.
% In addition, our results show that the accumulator is highly scalable.
% For example, adding adapters or increasing the depth of the accumulator will bring some improvement.
% Overall, our accumulator, which is both effective and scalable, is designed to take full advantage of FMs in FL.

% \input{tabs/ablation.tex}

%%%%%%%%%
% We can uncomment to put it back to main
% \paragraph{Communication Cost of Different Methods.}
% Table~\ref{tab:efficiency} shows the communication cost of different methods to reach a target test accuracy, which is set as the best performance of the L.W. Linear head here. We can see from the results that our \alg{} as such achieves the lowest cost to reach the target by only cost $40*3.17$ and $40*2.99$ message size for CIFAR-100 ($\alpha=0.1$) and FEMNIST(Non-IID).

% \input{tabs/efficiency.tex}
\section{Conclusions}
We explored the challenges and opportunities of leveraging pre-trained models for downstream federated learning tasks. While this might appear to raise communication costs and preclude diverse device capabilities from participating, we show that after  introducing a novel parameter-efficient adapter module we can  simultaneously capture the benefits of communication efficient adaptation, non-IID robustness, support for diverse device capabilities, and robust personalization.

\clearpage
\newpage
\bibliographystyle{apalike}
\bibliography{refs}

\clearpage
\newpage
\appendix

\section{Implementation Details}
\label{app:settings}
\paragraph{Hyperparameter Settings}
{The base setup is the same for all experiments, and we basically did not include any extra training tricks to be able to demonstrate that our proposed framework can be successfully adapted with \textbf{Anytime} and \textbf{Anywhere}. For all datasets, the initial learning rate for the SGD optimizer is set to 5e-3, and the batch size and training epoch on the client is set to 10 and 1, respectively. The \texttt{RandomResizedCrop} with scale=(0.05, 1.0) is used in training. The FedAvg~\cite{mcmahan2017communication} is the default aggregation algorithm.}

\paragraph{Conventional Federated Learning}
{In this setting, the last layer of the foundation model is set as the only exit. 
The full foundation model can be used for training the additional parameters, \textit{e.g.,} L.W. MLP, and Accumulator, \textit{etc}.
The baseline Fine-tuning indicates all parameters of the foundation model with an extra MLP head are free for tuning. 
The total training round is set to 500.}

\paragraph{Anytime Federated Learning}
{The major difference from the conventional FL is that the exit layer $l \in [1,12]$ for each client is randomly sampled from a uniform distribution for each training iteration.
The total training round is set to 1500.
}

\paragraph{Multi-tier based Federated Learning}
{Exit layer is no longer produced by random sampling, but a fixed value that represents the current computing power of the client.
In other words, each client will be assigned a permanent $l$ during the anytime FL. 
The distribution of clients' exit layer $l$ is balanced.
The total training round is set to 1500.
}

\paragraph{Personalization}
{Following the data corruption methods in ~\cite{hendrycks2019cifarC}, we apply a corruption policy with different severity $s \in [0, 5]$ as unique style for each client's local data.
Then, we fine-tune the pre-trained multi-tier model with a predefined exit layer on these data for 10 local epochs.}

\section{Additional Experiments}
\label{app:exps}
\paragraph{Communication Cost of Different Methods.}
Table~\ref{tab:efficiency} shows the communication cost of different methods to reach a target test accuracy, which is set as the best performance of the L.W. Linear head here. We can see from the results that our \alg{} as such achieves the lowest cost to reach the target by only cost $40*3.17$ and $40*2.99$ message size for CIFAR-100 ($\alpha=0.1$) and FEMNIST(Non-IID).
\begin{table}[t]
\caption{Transmitted message size (\# communication round$\times$\# model parameters (M)) required to reach a target performance with multi-tier FL on CIFAR-100 and FEMNIST (\textcolor{red}{best} and \textcolor{blue}{second best}).}
\centering
\scalebox{0.8}{
\begin{tabular}{l|c|c|c|c|c|c|c}
\toprule
& \multirow{2}{*}{Fine-tuning} & \multirow{2}{*}{L.W. Linear} & L.W. Linear         & \multirow{2}{*}{L.W. MLP} & L.W. MLP         & \multirow{2}{*}{Accumulator} & Accumulator \\ 
&             &        &  +Adapter      &     & +Adapter    &             & +Adapter \\ 
\midrule
CIFAR-100 & 90$\times$30.62 & 1500$\times$0.46 & 140$\times$1.06 & 610$\times$8.95 & 165$\times$9.55 & \textcolor{red}{40$\times$3.17} & \textcolor{blue}{100$\times$3.77} \\
\midrule
FEMNIST & 60$\times$30.44 & 1500$\times$0.28 & \textcolor{blue}{160$\times$0.88} & 1500$\times$8.77 & 240$\times$9.37 & \textcolor{red}{40$\times$2.99} & 80$\times$3.59 \\
\bottomrule
\end{tabular}
}
\label{tab:efficiency}
\end{table}

\begin{table}
\caption{{The computational and memory budgets across all exit layers with Deit-S as the foundation model. The memory and compute footprint increases linearly with the layer at which the early exit is triggered. During training, the memory peak can be largely reduced if the base model is kept frozen, allowing in this way the participation of more constrained devices.}}
\centering
\scalebox{0.8}{
\begin{tabular}{c|c|c|c|c|c|c|c|c|c|c|c|c}
\toprule
Early Exit Layer       & 0      & 1      & 2      & 3       & 4       & 5       & 6       & 7       & 8      & 9       & 10      & 11      \\ \midrule
FLOPs(GB)   & 0.21 & 0.40 & 0.57 & 0.67  & 0.95  & 1.13  & 1.31  & 1.50  & 1.69 & 1.88  & 2.07  & 2.67  \\ \midrule
Params(MB) & 2.54 & 4.34 & 6.15 & 7.96 & 9.77 & 11.57 & 13.38 &  15.19 & 16.99 & 18.80 & 20.60 & 22.41 \\
\midrule
Mem. Peak(MB) & 67  & 123 & 179 & 236 & 293 & 350  & 407 & 464 & 522 & 580 & 638 & 670\\
\midrule
Mem. Peak(MB)$_\text{frozen}$ & 47 & 62 & 98 & 134 & 170 & 206 & 242 & 279 & 315 & 352 & 389 & 426 \\

\bottomrule
\end{tabular}
}
\label{tab:budget}
\end{table}

\end{document}